\documentclass{article}

\usepackage[preprint]{spconf}
\usepackage{amsmath,graphicx}
\usepackage{multirow}
\usepackage{amssymb}
\usepackage{booktabs}
\usepackage{enumitem}
\usepackage{tikz}
\usepackage{hyperref}
\hypersetup{colorlinks=true}

\DeclareGraphicsExtensions{.pdf,.png}
\title{Partial Variable Training for Efficient On-Device Federated Learning}

\name{Tien-Ju Yang, Dhruv Guliani, Françoise Beaufays, Giovanni Motta}
\address{Google LLC, Mountain View, CA, U.S.A.\vspace{-5pt}}

\begin{document}

\maketitle


\newcommand\copyrighttext{
  \footnotesize \textcopyright 2021 IEEE. Personal use of this material is permitted. Permission from IEEE must be obtained for all other uses, in any current or future media, including reprinting/republishing this material for advertising or promotional purposes, creating new collective works, for resale or redistribution to servers or lists, or reuse of any copyrighted component of this work in other works.
}
\newcommand\copyrightnoticebottom{
    \begin{tikzpicture}[remember picture,overlay]
    \node[anchor=south,yshift=10pt] at (current page.south) {\fbox{\parbox{\dimexpr\textwidth-\fboxsep-\fboxrule\relax}{\copyrighttext}}};
    \end{tikzpicture}
}

\copyrightnoticebottom


\begin{abstract}
This paper aims to address the major challenges of Federated Learning (FL) on edge devices: limited memory and expensive communication. We propose a novel method, called \emph{Partial Variable Training (PVT)}, that only trains a small subset of variables on edge devices to reduce memory usage and communication cost. With PVT, we show that network accuracy can be maintained by utilizing more local training steps and devices, which is favorable for FL involving a large population of devices. According to our experiments on two state-of-the-art neural networks for speech recognition and two different datasets, PVT can reduce memory usage by up to 1.9$\times$ and communication cost by up to 593$\times$ while attaining comparable accuracy when compared with full network training.
\end{abstract}



\section{Introduction}
\label{sec:introduction}

Neural Networks (NNs) have become an indispensable backbone of many artificial intelligence applications, such as automatic speech recognition~\cite{graves_2013_asr}, that enrich our daily life. How well neural networks can serve users largely depends on the quality of data available for training. Edge devices generate a large amount of data constantly. However, the private nature of such data prevents them from being uploaded to servers to improve NNs and, hence, user experience.

Federated Learning (FL)~\cite{kairouz_2019_openproblem,wang_2021_fieldguide} is a promising solution to this dilemma by keeping data always on edge devices (referred to as \emph{clients}). FL starts from broadcasting a server network to clients and trains it on clients with local data. The resultant network changes are then sent back to the server and aggregated to update the server network. One set of these steps comprises a \emph{federated round}, and multiple rounds are carried out until the server network converges.

Performing FL on edge devices is challenging. In addition to the challenges the general FL faces~\cite{kairouz_2019_openproblem}, such as not independent and identically distributed (non-IID) data, on-device FL needs to address the efficiency problem. First, the memory is usually highly limited on edge devices. Edge devices can have as low as a few megabytes available for training. Second, communication is expensive. Communication can be orders of magnitude slower than local computation~\cite{huang_2013_communication_speed}.

In this paper, we propose \emph{Partial Variable Training (PVT)} to address the above challenges of on-device FL. On a client, the standard All-Variable Training (AVT) trains all the variables (e.g., weights and biases) and returns all the changes to the server. In contrast, PVT trains only a subset of variables and freezes the remaining per federated round. Only the changes corresponding to the trained variables are returned. The following summarizes the benefits of PVT:
\begin{itemize}[nolistsep]
    \item \textbf{Reducing memory usage:} Because memory usage is dominated by buffered activations for backward passes, freezing some variables prevents buffering the corresponding activations, hence it reduces memory usage.
    \item \textbf{Reducing communication cost:} Frozen variables are unchanged during training and do not need to be updated, thus reduced communication cost.
    \item \textbf{No need of modifying architectures:} PVT does not require inserting special layers or operations into networks and allows using the original architectures as is.
    \item \textbf{No requirement of network-specific knowledge:} PVT does not assume any special properties of a network or its layers, which makes PVT applicable to various networks.
    \item \textbf{Suiting large-scale federated learning:} We observe that the loss in accuracy when we freeze a large number of variables can be compensated by increasing the numbers of local training steps and clients. This trade-off is favorable for large-scale FL, where memory usage and communication cost are the hard constraints while there are many available devices.
\end{itemize}

\section{Methodology}
\label{sec:methodology}

\begin{figure}[t]
    \centering
    \includegraphics[width=0.35\textwidth]{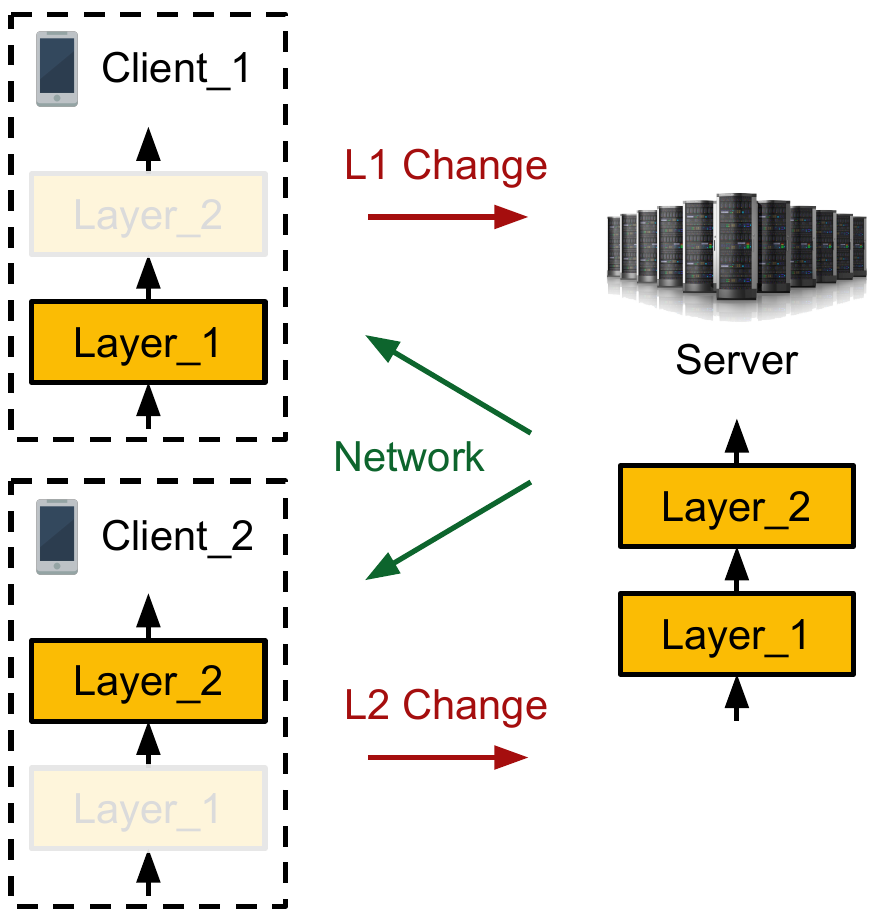}
    \vspace{-5pt}
    \caption{The illustration of the proposed partial variable training. In this example, $Client\_1$ only trains $Layer\_1$ and sends back its change. Similarly, $Client\_2$ only trains $Layer\_2$ and sends back its change. The server updates both layers after receiving the changes from both clients.}
    \label{fig:algorithm_illustration}
\end{figure}

Partial Variable Training (PVT) trains only a subset of variables and sends back the corresponding changes to a server. Fig.~\ref{fig:algorithm_illustration} illustrates one federated round of PVT. This simple example trains a two-layer network with two clients. On $Client\_1$, the variables in $Layer\_1$ are trained and those in $Layer\_2$ are frozen. On $Client\_2$, the variables in $Layer\_2$ are trained and those in $Layer\_1$ are frozen. The server then uses the change of $Layer\_1$ from $Client\_1$ and that of $Layer\_2$ from $Client\_2$ to update both layers of the server network. There are three main design decisions for PVT: what to freeze, how to freeze, and how many to freeze.

\subsection{What to Freeze}
\label{subsec:what_to_freeze}

PVT categorizes variables into two groups: freezable variables and non-freezable variables. PVT always trains the non-freezable variables and freezes a subset of freezable variables. To avoid the need for prior knowledge about the target network, the variable grouping is in the granularity of variables based on variable types. We define three types of variables:
\begin{itemize}
    \item \textbf{Additive vectors:} An additive vector adds a constant to each channel of input activations. Examples are the biases in convolutional layers and the offset factors in normalization layers. Additive vectors usually have a negligible cost in communication because they typically account for only a small portion of a network. Moreover, they also use less memory than the other variable types since fewer activations need to be buffered. In the example of updating biases ($b_{i}$) by gradients in a linear layer $a_{i+1} = a_{i} + b_{i}$, where $a_{*}$ represent activations, $\frac{\partial a_{i+1}}{\partial b_{i}}$ does not depend on $a_{i}$.
    \item \textbf{Multiplicative vectors:} A multiplicative vector scales each channel of input activations by a constant. Examples are the scaling factors in normalization layers. Similar to additive vectors, multiplicative vectors usually account for only a small portion of a network and have a negligible cost in communication. However, they require buffering activations for gradient computation, hence they consume more memory than additive vectors. In the example of updating scaling factors ($s_{i}$) by gradients in a linear layer $a_{i+1} = s_{i} * a_{i}$, $\frac{\partial a_{i+1}}{\partial s_{i}}$ is equal to $a_{i}$, so $a_{i}$ needs to be buffered.
    \item \textbf{Multiplicative matrices:} An input activation matrix is multiplied by a multiplicative matrix. Examples are the weight matrices in convolutional and feed-forward layers. They usually dominate the network size and require buffering activations for gradient computation. Hence, multiplicative matrices are the most expensive in terms of memory usage and communication cost.
\end{itemize}
Table~\ref{tab:variable_types} summarizes the memory usage and the communication cost of these three variable types. It is intuitive to freeze multiplicative matrices first, multiplicative vectors second, and additive vectors third. Moreover, prior knowledge on the target network, such as layer ambience~\cite{bengio_2019_ambientlayers}, can be incorporated to further improve the effectiveness of PVT.

\begin{table}[t]
\centering
\scalebox{0.9}{
\begin{tabular}{c|cc}
\toprule
                        & Memory & Communication \\ \hline
Additive Vectors        & Low    & Low           \\ 
Multiplicative Vectors  & High   & Low           \\ 
Multiplicative Matrices & High   & High          \\
\bottomrule
\end{tabular}
}
\caption{The memory usage and the communication cost of the three variable types.}
\label{tab:variable_types}
\end{table}

\subsection{How to Freeze}

The choice about which variables to freeze can
\begin{itemize}
    \item \textbf{Be fixed (fixed):} The same set of variables are chosen by all clients in all rounds.
    \item \textbf{Vary per round (PR):} The chosen variables vary from round to round, but all clients in a given round choose the same set of variables.
    \item \textbf{Vary per client per round (PCPR):} The chosen variables vary from round to round and from client to client.
\end{itemize}
PCPR generally attains the highest accuracy given the same number of rounds and works the best for from-scratch training because it updates all the variables in each round. Compared with PCPR, PR is easier to implement because all the clients share the same network graph at the cost of longer convergence time. The fixed scheme can be useful when we have prior knowledge about the target network.

\subsection{How Many to Freeze}

Freezing more variables enables higher memory and communication savings at the cost of accuracy degradation. Fortunately, we observe that almost all the loss in accuracy can be compensated by utilizing more local training steps and more clients. Therefore, by carefully choosing the hyperparameters (e.g., number of frozen variables, local training steps, and clients), PVT can improve FL efficiency while providing highly accurate networks.

We suggest the following steps to determine these hyper-parameters systematically. We first increase the number of frozen variables until the memory and communication constraints are satisfied. Then, we increase the number of local training steps to what is allowed on devices. Finally, we increase the number of clients until the accuracy is restored.

\section{Experimental Results}
\label{sec:experimental_results}

\subsection{Experimental Settings}
Two networks are adopted in the experiments to cover both non-streaming and streaming use cases. The first network is the largest Conformer~\cite{gulati_2020_conformer} (referred to as \emph{non-streaming Conformer}) with Batch Normalization replaced by Group Normalization, which is more suitable to federated learning~\cite{hsieh_2019_decentralized_ml} but at a small cost of accuracy. The second network is our production-grade Conformer variant~\cite{li_2021_streaming_conformer} (referred to as \emph{streaming Conformer}), which supports streaming use cases and contains approximately 137M trainable parameters.

In addition to two networks, we adopt two datasets with different partition methods to cover different data distributions and from-scratch training and domain adaptation scenarios. The first dataset is Librispeech~\cite{panayotov_2015_librispeech}. \emph{IID Librispeech} is generated by randomly partitioning Librispeech into multiple small datasets to simulate IID clients' local datasets. \emph{Non-IID Librispeech} is generated by partitioning Librispeech by speakers to simulate non-IID clients' local datasets. The two different partition methods help evaluate PVT under both IID and non-IID data. For experiments on Librispeech, we train networks from scratch with the training set and evaluate them on the test set.

The second dataset is an anonymized \emph{Multi-Domain (MD) dataset} containing approximately 400K hours of utterances from domains such as search,  farfield,  telephony, and YouTube~\cite{narayanan_2019_mddataset,misra_2021_mddataset}. We withhold one domain (\emph{Medium Form (MF)} in this paper) from the MD dataset to evaluate PVT under domain adaptation scenarios. The MF domain has approximately 26K hours of utterance with an average duration of 10.4 seconds. For experiments on the MD dataset, we first train networks on the training set with the MF domain withheld, refining them on the MF domain, and then evaluate them on a disjoint test set from the MF domain.

Unless otherwise specified, the non-freezable variables are all the additive vectors, and the freezable variables are all the multiplicative vectors and matrices. We randomly freeze 90\% of freezable variables with the PCPR scheme. There are 1024 clients, and each client trains a network with 5 local steps. The batch size is 16 per client. For resource consumption, we report the peak memory usage with gradient recomputation~\cite{chen_2016_gradient_recomputation} and the client-to-server (CtoS) communication cost. For Word Error Rates (WERs) on Librispeech, we report them in the format of \emph{dev/dev-other/test/test-other}, where each item corresponds to the WER of the dev, dev-other, test, and test-other set from left to right. 

\subsection{Non-Streaming Conformer on Librispeech}

Table~\ref{table:ns_cf_iid_libri} summarizes the results of non-streaming Conformer on IID Librispeech. Compared with all-variable training (AVT), PVT can achieve similar WERs with much lower memory usage and CtoS communication cost. We observe that running forward passes alone requires 761MB, and AVT requires extra 611MB to allow running backward passes. In contrast, PVT only requires extra 313MB, which is 1.9$\times$ lower than that of AVT. Moreover, PVT reduces the CtoS communication cost by 9.8$\times$.

\begin{table}[t]
\centering
\scalebox{0.9}{
\begin{tabular}{c|c|ccc}
\toprule
\multirow{2}{*}{} & \multirow{2}{*}{WERs} & \multicolumn{2}{c}{Resource} \\ \cline{3-4} 
                  &                       & Peak Memory  & CtoS Comm.   \\ \hline
FProp        & -                     & 761MB        & -    \\
AVT               & 2.1/4.6/2.2/4.8       & 1372MB        & 452MB    \\
\textbf{PVT}      & \textbf{2.1/4.9/2.3/4.8}      & \textbf{1074MB}        & \textbf{46MB}      \\
\bottomrule
\end{tabular}
}
\caption{The results of non-streaming Conformer on IID Librispeech. \emph{FProp} refers to performing only forward passes.}
\label{table:ns_cf_iid_libri}
\end{table}

Table~\ref{table:ns_cf_noniid_libri} summarizes the WERs of non-streaming Conformer on non-IID Librispeech. Even with non-IID data, PVT can still attain comparable WERs to AVT. The reduction in memory usage and CtoS communication cost is the same as the previous IID experiment and, hence, omitted in Table~\ref{table:ns_cf_noniid_libri}. These experiments show the versatility of PVT to work well with both IID and non-IID data.

\begin{table}[t]
\centering
\scalebox{0.9}{
\begin{tabular}{c|cc}
\toprule
    & AVT             & \textbf{PVT}             \\ \hline
WER & 2.0/4.7/2.2/4.9 & \textbf{2.1/4.8/2.2/5.0} \\
\bottomrule
\end{tabular}
}
\caption{The WERs of non-streaming Conformer on non-IID Librispeech.}
\label{table:ns_cf_noniid_libri}
\end{table}

\subsection{Streaming Conformer on Multi-Domain Dataset}

We observe that domain adaptation may allow training fewer variables than from-scratch training. In this experiment, only training biases (i.e., freezing all freezable variables) is enough for providing high accuracy. Table~\ref{table:s_cf_vs} summarizes the results of streaming Conformer on the multi-domain dataset. Compared to AVT, PVT can achieve similar WERs with 1.3$\times$ reduction in extra memory usage on top of running forward passes and 593$\times$ reduction in communication cost. This experiment demonstrates the effectiveness of PVT on domain adaptation and large-scale datasets.

\begin{table}[t]
\centering
\scalebox{0.9}{
\begin{tabular}{c|c|ccc}
\toprule
\multirow{2}{*}{} & \multirow{2}{*}{WER} & \multicolumn{2}{c}{Resource} \\ \cline{3-4} 
                  &                       & Peak Memory  & CtoS Comm.    \\ \hline
Before Refinement & 6.9       & -        & -           \\
FProp             & -         & 836MB    & -           \\
AVT               & 4.5       & 1606MB        & 522MB           \\
\textbf{PVT}      & \textbf{4.6} & \textbf{1448MB}   & \textbf{0.9MB}               \\
\bottomrule
\end{tabular}
}
\caption{The results of streaming Conformer on the multi-domain dataset. The WER is on the MF domain. \emph{FProp} refers to performing only forward passes.}
\vspace{-5pt}
\label{table:s_cf_vs}
\end{table}

\subsection{Ablation Study}

We use non-streaming Conformer on IID Librispeech to study the influence of different design decisions. We use 1 local training step per round and 128 clients in this section.

\subsubsection{Per-Round Scheme vs. Per-Client-Per-Round Scheme}

We observe that per-round (PR) scheme is unable to converge to reasonable WERs within 200K rounds while the per-client-per-round (PCPR) scheme attains 2.4/5.7/2.5/5.7, which are close to that of AVT. Our hypothesis is that PR only updates 10\% of the freezable variables per round, so needs much more rounds than PCPR to properly update the entire network.

\subsubsection{Number of Local Training Steps}

Table~\ref{table:num_local_steps} compares the WERs and the required numbers of rounds when various numbers of local training steps are used. We observe that using more local training steps improves WERs and speeds up convergence. Moreover, in this experiment, PVT allows as many as 128 local training steps and still provides converged results.

\begin{table}[t]
\centering
\scalebox{0.9}{
\begin{tabular}{c|cc}
\toprule
Number of Local Steps & WER & Number of Rounds \\ \hline 
1            & 2.2/5.5/2.4/5.4    & 350K          \\ 
5           & 2.1/4.9/2.2/4.9    & 155K          \\ 
128           & 2.0/4.8/2.2/4.6    & 15K          \\
\bottomrule
\end{tabular}
}
\caption{The comparison among various numbers of local steps with non-streaming Conformer on IID Librispeech.}
\label{table:num_local_steps}
\end{table}

\subsubsection{Number of Clients}

Table~\ref{table:num_clients} compares the WERs and the required numbers of rounds when various numbers of clients are used. Similar to numbers of local training steps, WERs and convergence speed improve as we use more clients. This study shows that the loss of WERs and convergence speed caused by freezing variables can be compensated by using more clients, which is a favorable property for large-scale federated learning.

\begin{table}[t]
\centering
\scalebox{0.9}{
\begin{tabular}{c|cc}
\toprule
Number of Clients & WER & Number of Rounds \\ \hline 
128        & 2.2/5.5/2.4/5.4    & 350K          \\ 
1024       & 2.1/5.0/2.3/4.9    & 205K          \\ 
2048       & 2.1/5.1/2.2/5.0    &  90K          \\ 
4096       & 2.1/5.1/2.2/5.0    &  60K          \\ 
\bottomrule
\end{tabular}
}
\caption{The comparison among various numbers of clients with non-streaming Conformer on IID Librispeech.}
\vspace{-7pt}
\label{table:num_clients}
\end{table}

\subsubsection{WER and Convergence Speed Improvement}

\begin{figure}[t]
    \centering
    \includegraphics[width=0.45\textwidth]{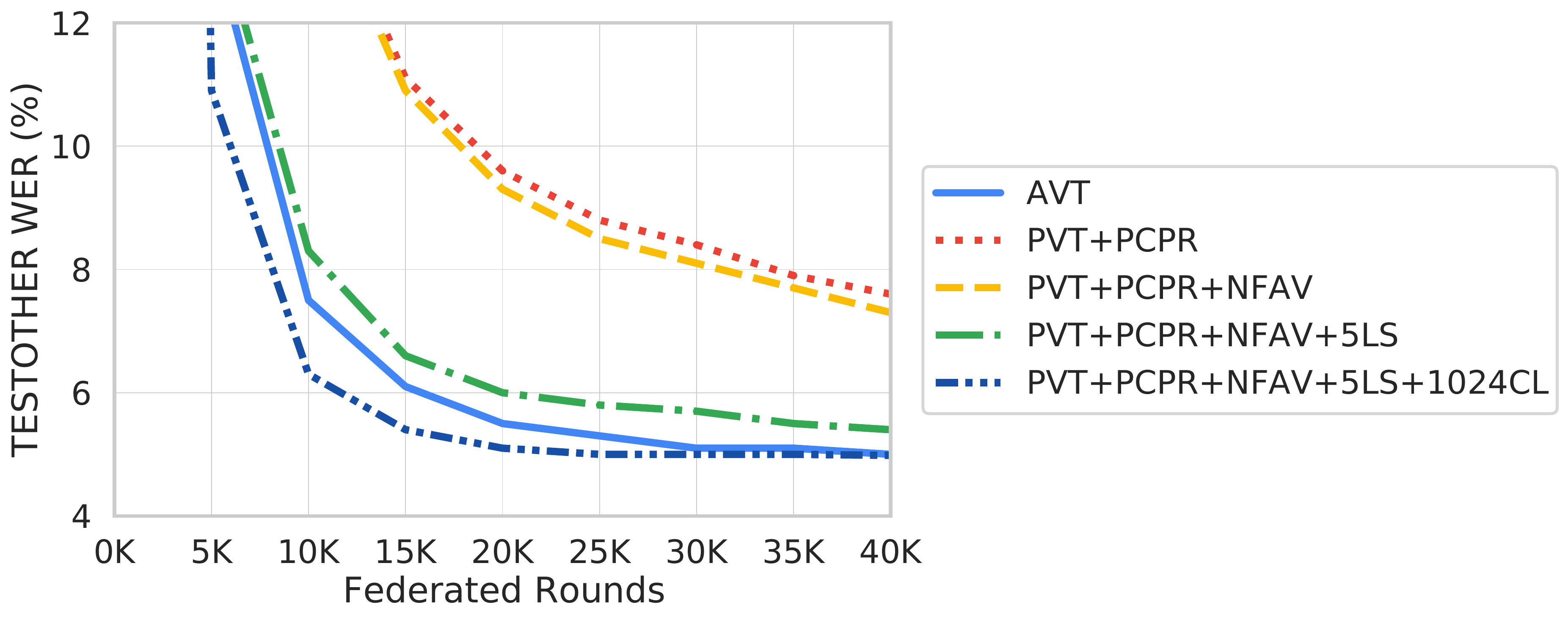}
    \caption{The WER and convergence speed improvement while we sequentially apply PCPR scheme, not freezing additive vectors (NFAV), 5 local steps (5LS), 1024 clients (1024CL) to train non-streaming Conformer on IID Librispeech.}
    \vspace{-1pt}
    \label{fig:wer_improvement}
\end{figure}

Fig.~\ref{fig:wer_improvement} shows how WERs and convergence speed improve while we sequentially apply what we learned from the analysis and ablation study above. When PCPR scheme, not-freezing additive vectors, 5 local training steps, and 1024 clients are sequentially applied, the training curve of PVT gradually becomes comparable to that of AVT.

\section{Related Works}
\label{sec:related_works}

There are strong efforts to tackle similar challenges as PVT in the literature. Model transport compression~\cite{chraibi_2019_model_broadcast_compression} and gradient transport compression~\cite{konecny_2016_gradient_compression} apply compression techniques to decrease the server-to-client and the client-to-server communication cost respectively, but the memory usage on devices is unchanged. Network compression reduces both memory usage and communication cost by simplifying networks, including manual design~\cite{sandler_2018_mobilenetv2}, pruning~\cite{yang_2017_energy_pruning}, quantization~\cite{rastegari_2016_xnornet}, and neural architecture search~\cite{yang_2021_netadaptv2}. However, reducing network complexity may degrade the ability of federated learning to benefit from a large amount of data, which requires high-complexity networks~\cite{brown_2020_gpt3}. Federated dropout ~\cite{caldas_2018_feddrop} addresses this issue by using simplified server networks on clients, so the server network can be complex without burdening clients. As server and client networks differ, federated dropout needs additional infrastructure to maintain a mapping of client networks to the full server network. Similar to federate dropout, group knowledge transfer~\cite{he_2020_knowledge_transfer} also uses different networks on a server and clients. The simpler client networks are used to extract the features for training the server network on a server, which decreases client loading at the cost of increased server loading. Compared to the above methods, PVT can reduce both memory usage and communication cost without their downsides and can be further combined with them to achieve even better efficiency.

\section{Conclusion}
\label{sec:conclusion}

In this paper, we proposed Partial Variable Training that significantly reduces memory usage and communication cost for Federated Learning (FL) with a negligible impact on accuracy. It does not require modifying architectures and network-specific knowledge and suits large-scale FL. We hope this technique will help bring FL onto edge devices and run in a large scale to greatly improve user experience.

\section{Acknowledgements}
\label{sec:acknowledgements}

We thank Yonghui Xiao, Petr Zadrazil, and Changwan Ryu for supporting memory measurement in this paper.


{
\bibliographystyle{IEEEbib}

}

\end{document}